\documentclass[letterpaper, 10 pt, journal, twoside]{ieeetran}

\IEEEoverridecommandlockouts                              

\usepackage{amsmath} 
\usepackage{amssymb}  
\usepackage{graphicx}
\usepackage{subcaption}
\usepackage{float}
\usepackage{amsmath}

\usepackage{pifont}
\usepackage{xcolor}
\usepackage{soul} 
\usepackage[colorlinks,
linkcolor=blue,
 anchorcolor=blue,
 citecolor=blue]{hyperref}
 \usepackage{overpic}
 \usepackage[font=small,labelfont=bf]{caption}

\title{
SRL-VIC: A Variable Stiffness-based Safe Reinforcement Learning \\for Contact-rich Robotic Tasks
}

\author{Heng Zhang$^{1,2}$, Gokhan Solak$^{1}$, Gustavo J. G. Lahr$^{1}$, Arash Ajoudani$^{1}$
\thanks{Manuscript received: January 11, 2024; Revised March 28, 2024; Accepted April 24, 2024.}
\thanks{This paper was recommended for publication by Editor Jens Kober upon evaluation of the Associate Editor and Reviewers' comments.
This work was supported part by the European Union’s Horizon 2020 research and innovation
programme SOPHIA (GA 871237) and CONCERT (GA 101016007).} 
\thanks{$^{1}$~Human-Robot Interfaces and Interaction Lab, Istituto Italiano di Tecnologia, Genoa, Italy.
        \mbox{e-mails: {heng.zhang@iit.it}}}%
\thanks{$^{2}$~Ph.D. program of national interest in Robotics and Intelligent Machines (DRIM) and Università di Genova, Genoa, Italy.}
\thanks{Digital Object Identifier (DOI): see top of this page.}
}

\begin{document}

\markboth{IEEE Robotics and Automation Letters. Preprint Version. Accepted Apr., 2024}
{Zhang \MakeLowercase{\textit{et al.}}: SRL-VIC: A Variable Stiffness-based Safe Reinforcement Learning for Contact-rich Robotic Tasks} 

\maketitle

\begin{abstract}

Reinforcement learning (RL) has emerged as a promising paradigm in complex and continuous robotic tasks, however, safe exploration has been one of the main challenges, especially in contact-rich manipulation tasks in unstructured environments. Focusing on this issue, we propose \textbf{SRL-VIC}: a model-free \textbf{s}afe \textbf{RL} framework combined with a variable impedance controller (\textbf{VIC}). Specifically, \textit{safety critic} and \textit{recovery policy} networks are pre-trained where \textit{safety critic} evaluates the safety of the next action using a risk value before it is executed and the \textit{recovery policy} suggests a corrective action if the risk value is high. 
Furthermore, the policies are updated online where the task policy not only achieves the task but also modulates the stiffness parameters to keep a safe and compliant profile.
A set of experiments in contact-rich maze tasks demonstrate that our framework outperforms the baselines (without the recovery mechanism and without the VIC), yielding a good trade-off between efficient task accomplishment and safety guarantee. 
We show our policy trained on simulation can be deployed on a physical robot without fine-tuning, achieving successful task completion with robustness and generalization. The video is available at \href{https://youtu.be/ksWXR3vByoQ}{https://youtu.be/ksWXR3vByoQ}.

\end{abstract}


\section{INTRODUCTION} \label{sec:intro}


\IEEEPARstart{E}{xploring} unstructured environments is a key skill for future robotics, where the robots will be employed in unknown and potentially hostile surroundings. The inherent physical interactions in such scenarios pose risks to robots, humans, and the environment. These risks are increased in the context of automated learning and exploration, e.g., in RL. Indeed, ensuring the safety of RL in tasks involving rich contact remains an ongoing challenge, especially in unstructured environments.

\begin{figure}[tb] 
      \centering  
      \includegraphics[width=0.9 \linewidth]{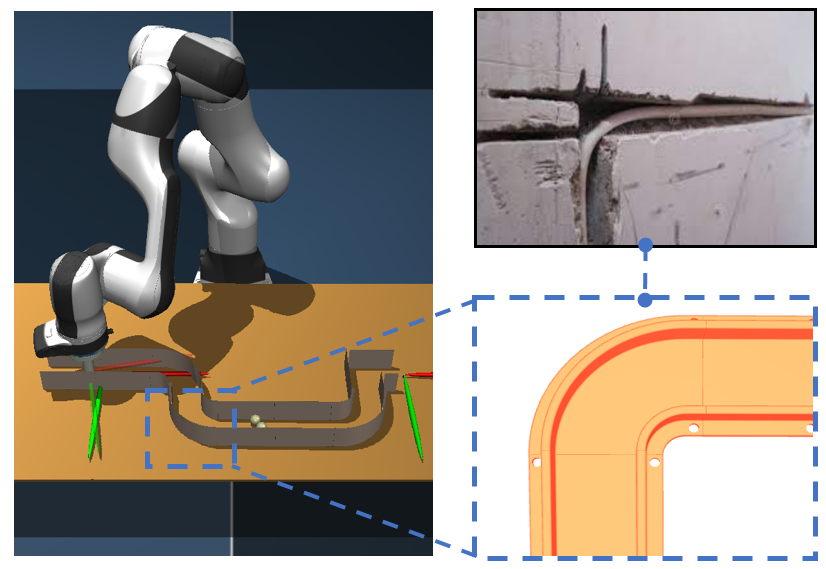} 
      \vspace{-0.1cm}
      \caption{Maze exploration is an unstructured contact-rich task that has practical applications such as cable-lying in walls. It also has similarities to search missions in dark and narrow environments. The agent does not have access to vision, thus it should navigate using only the contact information.}
      \label{fig:motivation} 
      \vspace{-0.3cm}
\end{figure}

We study the problem of safe RL on a maze exploration scenario as an illustrative example of contact-rich tasks. This scenario is inspired by the common construction task of installing electricity cables within walls, as depicted in Fig.~\ref{fig:motivation}. Traditionally, these tasks are done by workers and are time-consuming and labor-intensive. These characteristics indicate high potential for robotic automation, but raise several challenges: {1)} The environment is unknown to the robot; {2)} Visual perception may be difficult to deploy, thus physical contact is essential; {3)} Frequent contact with the environment causes safety issues.
To this end, we abstracted the cable-lying task into a robot exploration task since all the slots in the wall are composed of two basic shapes (turns and straight sections) as shown in Fig.~\ref{fig:motivation}. Then, we created an experimental setup that includes a 7-DoF robot arm and a maze with obstacles as shown in Fig.~\ref{fig:motivation}.

In tasks involving rich contact, the robot must engage in continuous interaction with the environment, presenting a notable safety risk that could result in damage to either the robot itself or objects in its surroundings. Impedance control is widely used to alleviate such hazards during physical robot interaction~\cite{ajoudani2018progress}.  Moreover, variable impedance control (VIC) methods enable adapting impedance parameters to the changing conditions of the task or the surrounding environment \cite{abu2020variable,kato2022self}. The robot can trade between compliance and accuracy to maximize task performance while staying safe.  Nevertheless, determining the adjustment of these parameters in VIC, particularly in unstructured environments with dynamic obstacles, remains a challenging question.

Learning-based approaches became a powerful way to address this challenge, with a wide variety of methods for tuning impedance parameters~\cite{abu2020variable}. In previous works, such parameters were learned from human (or expert) demonstrations using Dynamic movement primitives (DMPs)~\cite{chang2022impedance}, inverse reinforcement learning (IRL)~\cite{zhang2021learning} and skill priors~\cite{yang2022variable}; however, providing demonstrations in advance is relatively time-consuming and labor-intensive. 
Also, it can be impractical to provide expert demonstrations in some tasks where the environment is unstructured or potentially hostile.


In summarizing the current state-of-the-art research on contact-rich tasks, two notable approaches emerge: 
{1)} Some studies employ VIC to ensure safe performance while utilizing RL for learning impedance parameters \cite{yang2022variable}\cite{martin2019variable}; 
{2)} Other works leverage the predictive capabilities of neural networks to anticipate the safety value of future actions, allowing for informed decisions on whether to pursue a safer course of action in advance \cite{thananjeyan2021recovery}\cite{kim2023safety}. 
Each of these ideas has its own advantage: the former is to adjust the behaviour of the robot reactively based on the observed contact forces, while the latter is to proactively select a safer robot behaviour by predicting the safety of future actions, therefore avoiding actions that would generate dangerous amounts of contact force. They both improve the safety of robots when performing contact-rich tasks to some extent. 

Compared with other previous work, in this paper, we take a step towards combining the advantages of an RL-based VIC approach with the advantages of using a predictive safe RL approach. We introduce a new safe-RL framework incorporating VIC for contact-rich tasks. 
Following the recovery RL (RRL) method~\cite{thananjeyan2021recovery}, during the pre-training phase, we train a \textit{Safety-Critic} network using procedurally collected offline data, which can predict a risk value based on the current state and the next action. 
Meanwhile, we pre-train a \textit{recovery policy} network, which can sample a safe action if the risk value exceeds the safety threshold. In the online training phase, a task policy is trained with the goal of maximizing the reward. Unlike previous works, we do not use expert demonstrations to accelerate task policy training, given that the environment is unknown to the robot.

We compare the proposed framework with scenarios where VIC is absent and where a recovery mechanism is not implemented. The results of our simulation experiments indicate that the combination of VIC and recovery mechanism improves the safety and efficiency of the solution.


To summarize, the main contributions of our work are listed as follows: 

1) We present a safe RL framework with VIC for contact-rich tasks without prior knowledge of the environment. Prior to action execution, the \textit{safety critic}  predicts the safety of the action. The \textit{recovery policy} is activated for risky situations. During the action execution, the contact force is regulated with the help of VIC to enhance safety.

2) We developed a contact-rich maze exploration task in simulations, and empirically demonstrated the proposed framework outperforms baselines. 

3) We demonstrate that our policy trained on simulation can be deployed on a physical robot without fine-tuning, successfully achieving the tasks even with different type of obstacles, maze sizes and shapes, and  flange sizes. 

\section{Related work}
\subsection{Contact-rich robotic tasks}
The solutions for contact-rich tasks have undergone a shift from contact modeling analysis \cite{qiao1993robotic} to learning-based approaches \cite{lee2020making} \cite{schoettler2020meta}. However, safety issues have always been an inescapable aspect in such contact-rich tasks.

\subsection{Variable impedance control for contact-rich tasks}
Implementing VIC enables a robot to exhibit a certain degree of compliance, contributing to the achievement of safe behavior in contact-rich tasks. However, it is a challenge to pre-define the impedance parameters according to different tasks or different stages of the task. To solve this problem, Yang et al. \cite{yang2015teaching} utilized muscle surface electromyography (sEMG) signals to extract the demonstrator's variable stiffness which were passed to the robot. In this type of method, however, specialized equipment is required. Lee et al. \cite{7041441} proposed a stiffness modulation method that achieved safe and robust performance in a door opening task, while it focused on stiffness in joint space. A self-turning impedance was proposed in \cite{kato2022self}, which ensures compliant contact and low contact forces based on two planning strategies. However, the direction of motion of the robot will not change unless large contact forces are generated, which means that the actual trajectory executed by the robot may not be optimal with respect to energy and time. Furthermore, some scholars \cite{yang2022variable,mitrovic2011learning,li2018force} learn different impedance parameters from collected expert demonstration trajectories. However, it is difficult to demonstrate trajectories for tasks like ours, i.e., blind maze exploration. Moreover, some works address it by utilizing the impedance properties of the human body. Yang et al.~\cite{yang2015teaching,yang2018dmps} transferred the stiffness profile from the upper limb of the human to a humanoid dual-arm robot for contact-rich tasks. 

Therefore, these methods of obtaining variable impedance parameters either need demonstrations or require special devices or are specific to a single task.
\subsection{Safe RL for contact-rich tasks}
RL-based works focus on contact-rich tasks can be divided into two groups depending on when contact safety is increased. On the one hand, a safe controller with variable stiffness can be learned via RL. Specifically, Martin-Martin et al. \cite{martin2019variable} compared the effect of different action spaces in RL with variable impedance control in end-effector space for contact-rich tasks, such as surface-wiping and door-opening. 
Some work \cite{chang2022impedance,zhang2021learning,yang2022variable} leveraged demonstrations to learn variable impedance parameters as part of the action space. On the other hand, a safe \textit{action} or \textit{policy} can be learned with RL to significantly enhance safety during the training stage. For instance, MoPA-RL \cite{yamada2021motion} augmented the action space with long-horizon planning to improve safety. \textit{Safety critic} network can be trained that generate a risk value to help task policy sample a conservative or relatively safe action, works such as RRL \cite{thananjeyan2021recovery} and CSC \cite{bharadhwaj2020conservative}.

\begin{figure}[tb] 
      \centering  
      \includegraphics[width=0.86\columnwidth]{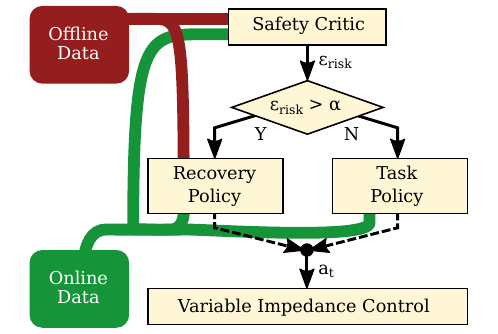} 
      \caption{The proposed framework: we combine the recovery-based safe RL approach \cite{thananjeyan2021recovery} with VIC to solve the contact-rich maze exploration task. 
      We first use an automated procedure to collect the offline data and pre-train our safety critic and recovery policy. Then, we train all learning components using online data. 
      The risk value $\epsilon_{risk}$ is used to activate either the task or recovery policy. 
      The action $a_t$ is chosen by the activated component, and it is fed to the VIC. Our action includes a relative position change and desired stiffness vector $\{K_x, K_y\}$.}
      \label{fig:framework} 
\end{figure}

However, in contrast to these, our proposed framework combines the strengths of both. We improve safety in terms of both impedance parameters in the controller and safety prediction during training. Furthermore, unlike most of the existing methods, we do not use expert demonstrations for task learning. 
Our method uses a scripted procedure to demonstrate only the concept of collision constraints which is a more immediate and simpler concept compared to a long-term task, hence easier to program. 

\section{Approach}
We introduce the details of our approach in this section. In Section~\ref{subsection:VIC}, the fundamentals of impedance control in task space are presented, which is our downstream controller. Then, we describe the proposed RRL-based VIC in Section~\ref{subsection:SafeRL}. A diagram of the proposed framework is shown in Fig.~\ref{fig:framework}. 

\subsection{Impedance control in task space} \label{subsection:VIC}

The robotic manipulator rigid body model in joint space may be written as
\begin{equation} \label{eq:rigid_body_full_eq}
    \boldsymbol{M}(\boldsymbol{q}) \ddot{\boldsymbol{q}} + \boldsymbol{C}(\boldsymbol{q},\dot{\boldsymbol{q}})\dot{\boldsymbol{q}} + \boldsymbol{g}(\boldsymbol{q}) = \boldsymbol{\tau} + \boldsymbol{\tau}_{ext},
\end{equation}
where $\boldsymbol{q}, \dot{\boldsymbol{q}}, \ddot{\boldsymbol{q}} \in \mathbb{R}^n $ are the position, velocity, and acceleration vectors in joint space, respectively, and $n$ is the number of DoF. $\boldsymbol{M}(\boldsymbol{q}) \in \mathbb{R}^{n \times n}$ is the joint space inertia matrix, $\boldsymbol{C}(\boldsymbol{q}, \dot{\boldsymbol{q}}) \in \mathbb{R}^{n}$ the Coriolis term, $\boldsymbol{g}(\boldsymbol{q})\in \mathbb{R}^{n}$ the gravity term, and $\boldsymbol{\tau} \in \mathbb{R}^{n}$ is the actuation torques. The external torques generated by the interaction forces are given by $\boldsymbol{\tau}_{ext} = \boldsymbol{J}^T(\boldsymbol{q}) \boldsymbol{F}_{ext}$, being $\boldsymbol{J}^T(\boldsymbol{q}) \in \mathbb{R}^{n \times 6}$ the Jacobian transpose matrix.

The actuation torques are chosen to compensate for the Coriolis and gravitational vectors, plus to deal with the external forces risen by contact with the environment through an impedance model in task space:

\begin{equation}
\boldsymbol{F}_{ext} =  \boldsymbol{D}^d\dot{\Tilde{\boldsymbol{x}}} + \boldsymbol{K}^d\Tilde{\boldsymbol{x}},
    \label{eq:cartesian_impedance}
\end{equation}
where $\Tilde{\boldsymbol{x}}$ is the Cartesian pose error given by the difference between the desired Cartesian pose $\boldsymbol{x}_d\in\mathbb{R}^{6}$ and the actual pose $\boldsymbol{x}\in\mathbb{R}^{6}$. Accordingly, $\dot{\Tilde{\boldsymbol{x}}}$ is the velocity error between the desired and actual end-effector's velocity, $\dot{\boldsymbol{x}}, \dot{\boldsymbol{x}}_d\in\mathbb{R}^{6}$, respectively. The desired stiffness matrix $\boldsymbol{K}^d \in \mathbb{R}^{6 \times 6}$ is a diagonal matrix with variable terms as $\boldsymbol{K}^d$=$diag\{K_x, K_y, K_z, 0.15K_x, 0.15K_y, 0\}$, where the rotation elements for $[K_x, K_y]$ are set as 0.15 times of $[K_x, K_y]$ inspired by~\cite{kato2022self} to reduce and simplify the parameters tuning. We set the rotational z-axis stiffness to zero, to keep it indifferent w.r.t. the end-effector roll. The desired Cartesian damping matrix $\boldsymbol{D}^d \in \mathbb{R}^{6 \times 6}$ is also diagonal and its terms are defined by the double diagonal rule, i.e., $\boldsymbol{D}_i^d(t) = 2 \zeta \sqrt{\boldsymbol{K}_i^d}$ where $i$ represents the linear or rotational components in Task Space. We use $\zeta{=}0.707$ in our experiments. 
The multi-dimensional stiffness can be represented with a stiffness ellipsoid having different compliance in different dimensions \cite{kato2022self}.



\subsection{Safe RL for contact-rich tasks} \label{subsection:SafeRL}
We adopt the RRL framework \cite{thananjeyan2021recovery} to actively predict safety hazards and take corrective actions. The original RRL was not specifically designed for contact-rich tasks. 
In~\cite{thananjeyan2021recovery} contact was avoided, on the contrary, in our work continuous contact is required but possibly unsafe. 
Thus, we introduce impedance control in the task space to better handle the contact-rich aspect of the problem. Specifically, we add the Cartesian impedance parameters to the action space in RL. 


As shown in Fig.~\ref{fig:framework}, there are three components in our framework: Firstly, similar to RRL, a \textit{safety critic} network and \textit{recovery policy} are trained with the pre-collected offline data. Then, our task policy was trained online with the maze exploration experiments. The output of the task policy is given to the variable impedance controller which controls the robot to achieve the desired actions.

In this framework, the problem is defined as a constrained Markov decision process (CMDP) problem \cite{altman1995constrained}.
A tuple $(\mathcal{S}, \mathcal{A}, R, P, \gamma, \mu, \mathcal{C})$ denotes a CMDP, where $\mathcal{S}$ is the state space, $ \mathcal{A}$ is the action space, $R: \mathcal{S} \times \mathcal{A} \rightarrow \mathbb{R}$ is the reward function, $P$ is the state transition probability, $\gamma \in(0,1)$ is a reward discount factor, $\mu$ is the starting state distribution and $\mathcal{C}=\left\{\left(c_i: \mathcal{S} \rightarrow\{0,1\}, \chi_i \in \mathbb{R}\right) \mid i \in \mathbb{Z}\right\}$ denote safety constraints that the agent must satisfy. 
In our task, we set $\gamma$ as 0.9 and we define the other components as detailed below. 

\begin{itemize}
    \item {State Space $\mathcal{S}$}. To make our framework more general to contact-rich exploration tasks instead of this specific maze, we differentiate the state space across different networks: it is 6-dimensional vector for \textit{safety critic} network and \textit{recovery policy} including 6 force/torque [$F_x, F_y, F_z, T_x, T_y, T_z$] values measured from a F/T sensor. We do not give the position as an input to the safety network, so that it does not memorise the locations of the walls. While for task policy, we add the position [$P_x, P_y, P_z$] of the end-effector, obtaining a 9-dimensional vector as state space so that the agent remembers the position when finding its way to the goal.
    
    \item {Action Space $\mathcal{A}$}. We combine 2 impedance parameters $[K_x, K_y]$ with 2 end-effector displacement values $[\Delta P_x, \Delta P_y]$ which consist of 4-dimensional vector as action space. Note that here the displacement values $\Delta P$ are not absolute positions in the world coordinate but the desired change w.r.t. the end-effector position. The sample range of $\Delta P$ is [-0.03, 0.03] \textit{m} and the impedance values $[K_x, K_y]$ are within the sample range of [300, 1000] \textit{N/m}. We project the stiffness values onto the $\Delta P$ vector components as $[K_x^*, K_y^*]$ in the controller: 
    \begin{equation}\label{equ:K_along_motion}
    \begin{aligned}
    K_x^* =& |K_x \cdot \Delta P_x / \left \|\Delta \boldsymbol{P} \right \|| + |K_y \cdot -\Delta P_y / \left \|\Delta \boldsymbol{P} \right \|| \\
    K_y^* =& |K_x \cdot \Delta P_y / \left \|\Delta \boldsymbol{P} \right \|| + |K_y \cdot \Delta P_x / \left \|\Delta \boldsymbol{P} \right \|| 
    \end{aligned}
    \end{equation}
    
    \item {Reward Function $R$}. We use a negative Euclidean distance between the current position $P_{cur}$ of the end-effector and the goal position $P_{goal}$ multiplied by a constant $c_{pos}{=}100$. We introduce a penalty for high-force collisions $r_{col}{=}{-}250$ and a penalty for leaving from the entrance $r_{ent}{=}{-}500$ to satisfy our task requirements. Also we add a bonus for reaching the goal $r_{goal}{=}1000$ to increase the speed of convergence. 

    \item {Safety Constraints $\mathcal{C}$}. The force constraint requires the measured force magnitude to stay below a threshold. 
\end{itemize}


\subsubsection{\textit{Safety critic} and \textit{recovery policy}}
\label{subsub:SafetyCriticRecoveryPolicy}
Based on \cite{srinivasan2020learning}, our \textit{safety critic} learn a critic function $Q^\pi_{risk}$:
\begin{subequations}\label{eq:qrisk}
\begin{alignat}{2}%
&Q_{\text{risk}}^\pi\left(s_t, a_t\right) =\mathbb{E}_\pi\left[\sum_{t^{\prime}=t}^{\infty} \gamma_{\text {risk }}^{t^{\prime}-t} c_{t^{\prime}} \mid s_t, a_t\right] \tag{\ref{eq:qrisk}} \\
&\quad=c_t+\left(1-c_t\right) \gamma_{\text {risk }} \mathbb{E}_\pi\left[Q_{\text {risk }}^\pi\left(s_{t+1}, a_{t+1}\right) \mid s_t, a_t\right] \nonumber
\end{alignat}
\end{subequations}
where $c_t = 1$ denotes constraint violation, and $c_t = 0$ indicates safety in state $s_t$. $Q^\pi_{risk}$ is trained by minimizing a MSE loss function:
\begin{subequations} \label{eq:mse}
\begin{alignat}{2}
& J_{\text {risk }}\left(s_t, a_t, s_{t+1} ; \phi\right)=\frac{1}{2}\biggl(\hat{Q}_{\phi,\text{risk}}^\pi\left(s_t, a_t\right)- \tag{\ref{eq:mse}} \\
&\left(c_t+\left(1-c_t\right) \gamma_{\text{risk}} \underset{a_{t+1} \sim \pi\left(\cdot \mid s_{t+1}\right)}{E} \left[\hat{Q}_{\phi, \text{risk}}^\pi\left(s_{t+1}, a_{t+1}\right)\right]\right)\biggr)^2  \nonumber
\end{alignat}
\end{subequations}

Similar to \cite{thananjeyan2021recovery}, we use an off-policy DDPG~\cite{lillicrap2015continuous} algorithm to learn \textit{recovery} policy $\pi_{rec}$ by performing gradient descent on the \textit{safety critic} network $\hat{Q}_{\phi, \text { risk }}^\pi\left(s, \pi_{\text {rec}}(s)\right)$.

\subsubsection{Pre-training}
By providing a set of offline data for the agent, we expect our agent to learn the relationship of safety with contact force and motion. 
It is possible to avoid catastrophic collisions caused by unexpected actions during online training by learning the implicit relationship between the two in advance. Details of the data collection procedure are described in Section \ref{sub:dataCollection}. 
Unlike online training where the agent searches for actions that maximise the reward, in this stage, we run a scripted behavior to collect examples of collisions without any reward. 
This approach reduces the problem complexity by separating task learning and constraint learning which are otherwise interlaced.

We do not pre-train the task policy, because this requires recording complex expert demonstrations which are costly and difficult to obtain for a blind maze exploration task.

We pre-train the \textit{safety critic} network and \textit{recovery} policy with the offline data which are collected in simulation. 
Specifically, We use 256 as batch size according to the number of hidden units in the network and update the \textit{safety critic} network and \textit{recovery} policy 10000 times in the pre-training stage. 

\subsubsection{Task policy}
The aim of the task policy is to finish the task and obtain maximum rewards within the constraints of safety. 
We train the task policy by the SAC algorithm \cite{haarnoja2018soft}, which is based on the idea of maximum entropy. The main difference from other RL algorithms is that SAC maximizes the entropy while optimizing the policy for higher cumulative returns. For our task, the benefit of introducing entropy into the learning stage is the increased randomization. Therefore, the agent can sufficiently explore the state space to avoid getting stuck in the local optimum.
Our task policy works in complementarity with the recovery network described in \ref{subsub:SafetyCriticRecoveryPolicy}. Action is selected as follows depending on the risk value:
\begin{equation}
a_t= \begin{cases}a_t^{\pi_{\text {task }}} & {Q}_{\left(s_t, a_t^{\pi_{\text {task }}}\right)} \leq \epsilon_{risk} \\ a_t^{\pi_{\mathrm{rec}}} & {Q}_{\left(s_t, a_t^{\pi_{\text {task }}}\right)} > \epsilon_{risk}
\end{cases}
\end{equation}
where $a_t^{\pi_{\text {task }}} \sim \pi_{\text {task }}\left(\cdot \mid s_t\right)$ and $a_t^{\pi_{\text {rec }}} \sim \pi_{\text {rec }}\left(\cdot \mid s_t\right)$. Different from task policy, the \textit{recovery} policy is trained to minimize the $\hat{Q}_{\phi, \text { risk }}^\pi\left(s_t, a_t\right)$ so that it can provide a safe action $a_t^{\pi_{\text {rec }}}$ with small risk value i.e. $Q_{\left(s_t, a_t^{\pi_{\text rec }}\right)} < \epsilon_{risk}$.
\subsubsection{Online training}
All of the policy and \textit{safety critic} networks are updated online, where task policy $\pi_{task}$ is only trained in the online RL stage while recovery policy $\pi_{rec}$ and \textit{safety critic} are also pre-trained with offline data.

In every single step of training, we update the replay buffer with online data where each one is denoted as a tuple $[s_t, a_t, r_t, s_{t+1}, done]$. 
We set 500 as horizon for each episode, in each transition, actions are sampled from the policies and are executed continuously until one of the termination conditions is met. 
The termination conditions are: violating a constraint, leaving the maze from the entrance, achieving success, and reaching the maximum step count.
The networks are updated online after collecting each observation.

\section{Evaluation}
In the following evaluation experiments, we aim to study whether the proposed framework is: {1)} safer; 2) more efficient, in comparison to our baselines, which are standard RL with VIC, and RRL with constant stiffness ($K{=}300$ and $1000$ \textit{N/m}). 
We label the baselines in this order as \texttt{Std\_RL-VIC}, \texttt{SRL-K300}, \texttt{SRL-K1000}. 
In \texttt{Std\_RL-VIC}, we use the same RL algorithm without the recovery mechanism to highlight the added benefit of our framework that combines VIC with RRL. The other baselines \texttt{SRL-K300} and \texttt{SRL-K1000} are used to evaluate the benefit of the VIC. 
Safety can be judged by the absence of excessively high interaction forces, whereas efficiency can be defined as attaining high task successes and reward with less training.

\subsection{Experiment setup} \label{sub:setup}
We evaluate the proposed framework on a maze exploration task using the 7-DoF Franka Emika Panda arm in Mujoco simulation environment as shown in Fig.~\ref{fig:motivation}.
We use a peg-shaped flange mounted on the end-effector of the robot with the diameter of 30 mm and the length of 55 mm. The maze channel is 50 mm in width 70.35 cm in length. Because there is a small space for the robot to explore width-wise, we can treat it also as an extended peg-in-hole task. The goal for the agent is to reach from entrance to exit without constraint violations. We adhere to safety level 1 \cite{brunke2022safe}, in which the agent will receive a penalty when it violates the constraints, and the current training episode will be terminated.
\footnotetext[3]{The orange circle denotes goal point of the task, the cyan circles and yellow arrows along the trajectory represent the stiffness ellipsoids and the motion directions respectively.\label{fn:traj}}

This problem combines two challenges of finding the way to exit without vision, and staying safe by avoiding high contact forces. 
It creates a dilemma as the contacts are absolutely necessary to navigate while they are also risky. 
Furthermore, we add movable obstacles in the maze that requires the robot to push harder.  
Thus, resorting to a conservative policy cannot solve the problem. 
The agent needs to learn both the safe behaviour and the task achievement simultaneously.

We implement the obstacles as three heavy balls in simulation, and a pile of bolts in real-world. We position the obstacles in the middle of the maze.


We choose MuJoCo simulator (version 2.3.3) for its favorable multi-contact physics computation, suitable for contact-rich tasks. However, this required us to split the maze mesh into smaller convex components because non-convex objects are not allowed in MuJoCo. We used V-HACD\footnotetext[4]{\label{fn:vacd}https://github.com/kmammou/v-hacd} library to decompose our maze into 128 convex primitive shapes and assembled them into the same maze.

\begin{figure}[t] 
      \centering
      \begin{overpic}[width=1\columnwidth]{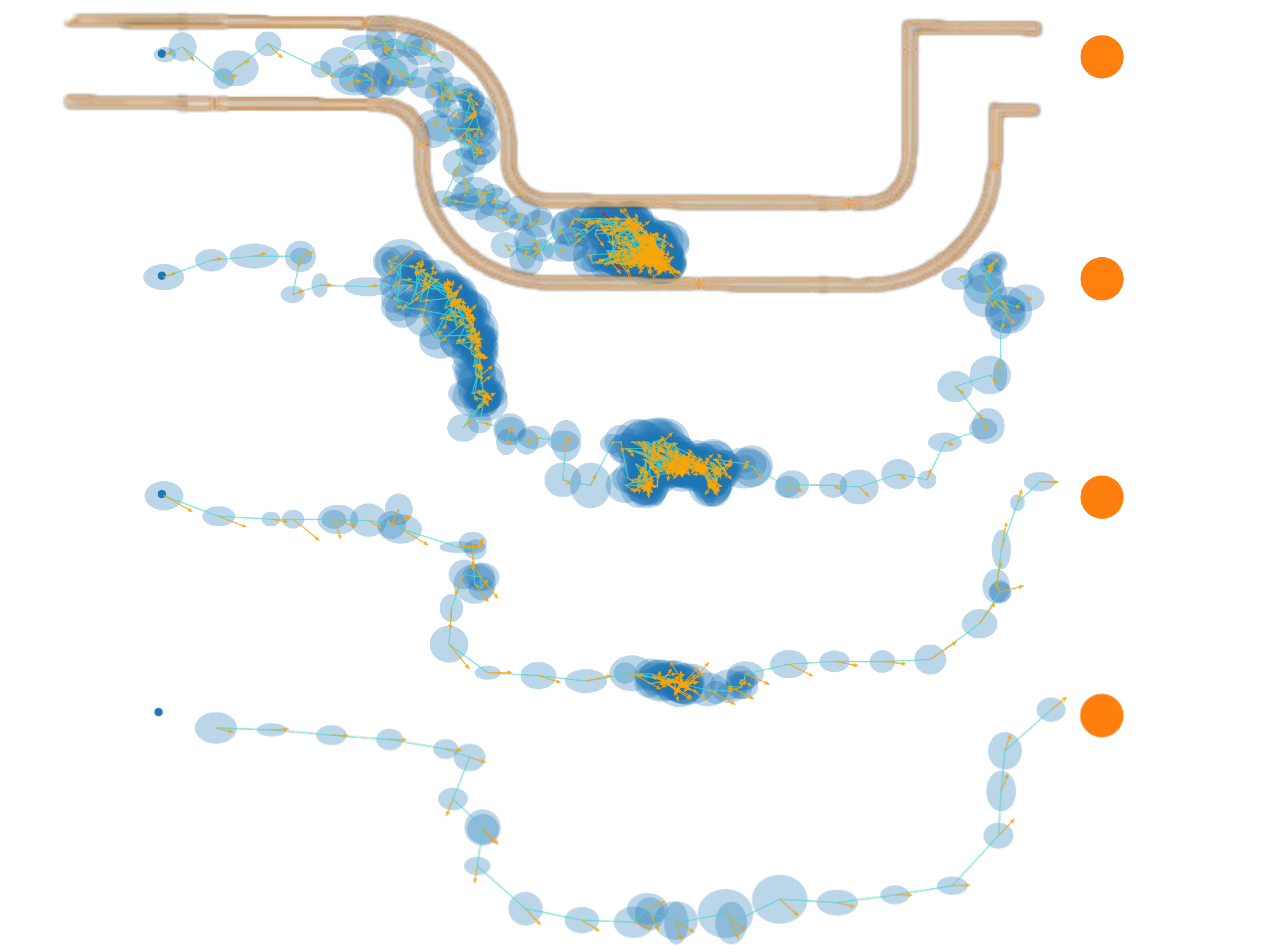}
        \put(4, 62){\small \bfseries \color{teal}{episode 10}}
        \put(4, 47){\small \bfseries \color{teal}{episode 40}}
        \put(4, 28){\small \bfseries \color{teal}{episode 210}}
        \put(4, 12){\small \bfseries \color{teal}{episode 1500}}
    \end{overpic}
      \vspace{-0.4cm}
      \caption[bla]{Trajectories of the end-effector during different stages of \textit{training}, in top-down view\textsuperscript{\ref{fn:traj}}.
      } 
      \vspace{-0.3cm}
      \label{fig:mulTraj} 
\end{figure}
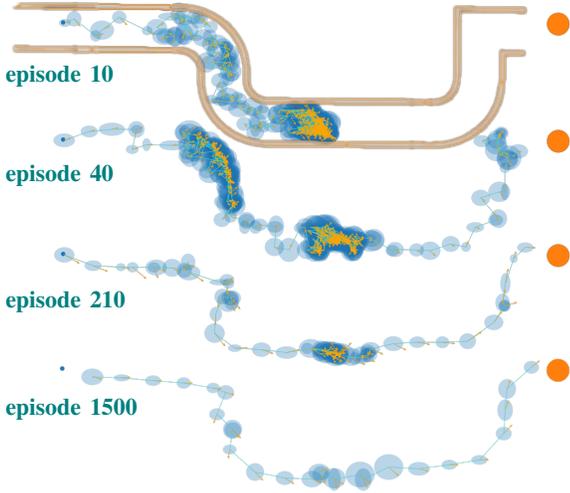 

The safe recovery RL part of our work is implemented based on the RRL codebase\footnote[5]{\label{fn:rrl}https://github.com/abalakrishna123/recovery-rl}. We follow the same hyperparameter procedure, except for: $10$ training updates per simulation step, 0.65 as $\gamma\_risk$, 0.7 as $\epsilon\_risk$ and 500 as the maximum number of actions for each episode. 
Another difference from RRL is that we normalize the actions of the RL model and denormalize them before applying them in the robot controller, for faster convergence. 

\subsection{Simulation experiments}

We first collect offline data to pre-train the safety critic and the recovery policy. Then, we train all approaches online and compare their performances.

\subsubsection{Offline data collection} \label{sub:dataCollection}
In order to teach the notion of risk, we generate random states and actions that are likely to cause constraint violation. 
The robot moves to one of the six predefined points inside the maze, then moves in a random direction, which is likely to collide with the maze walls. The number of transitions are split equally between each point. 
In order to make the collected data more diverse, we add noise to the coordinates of each point, so that we have many different starting positions. Then, we sample a \textit{move direction} as an angle between $0^\circ$ and $360^\circ$ and a \textit{move size} in the action range. 
Lastly, the robot is commanded to move in that direction and the transition is stored as a tuple $(s_t, a_t, c_t, s_{t+1}, done)$. Offline data for pretraining are given $40018$ transitions of data, $185$ of these transitions contain constraint-violating states.

\begin{figure}[tb]
    \centering
    \includegraphics[trim=2.5cm 2.5cm 6cm 1cm,width=1 \linewidth]{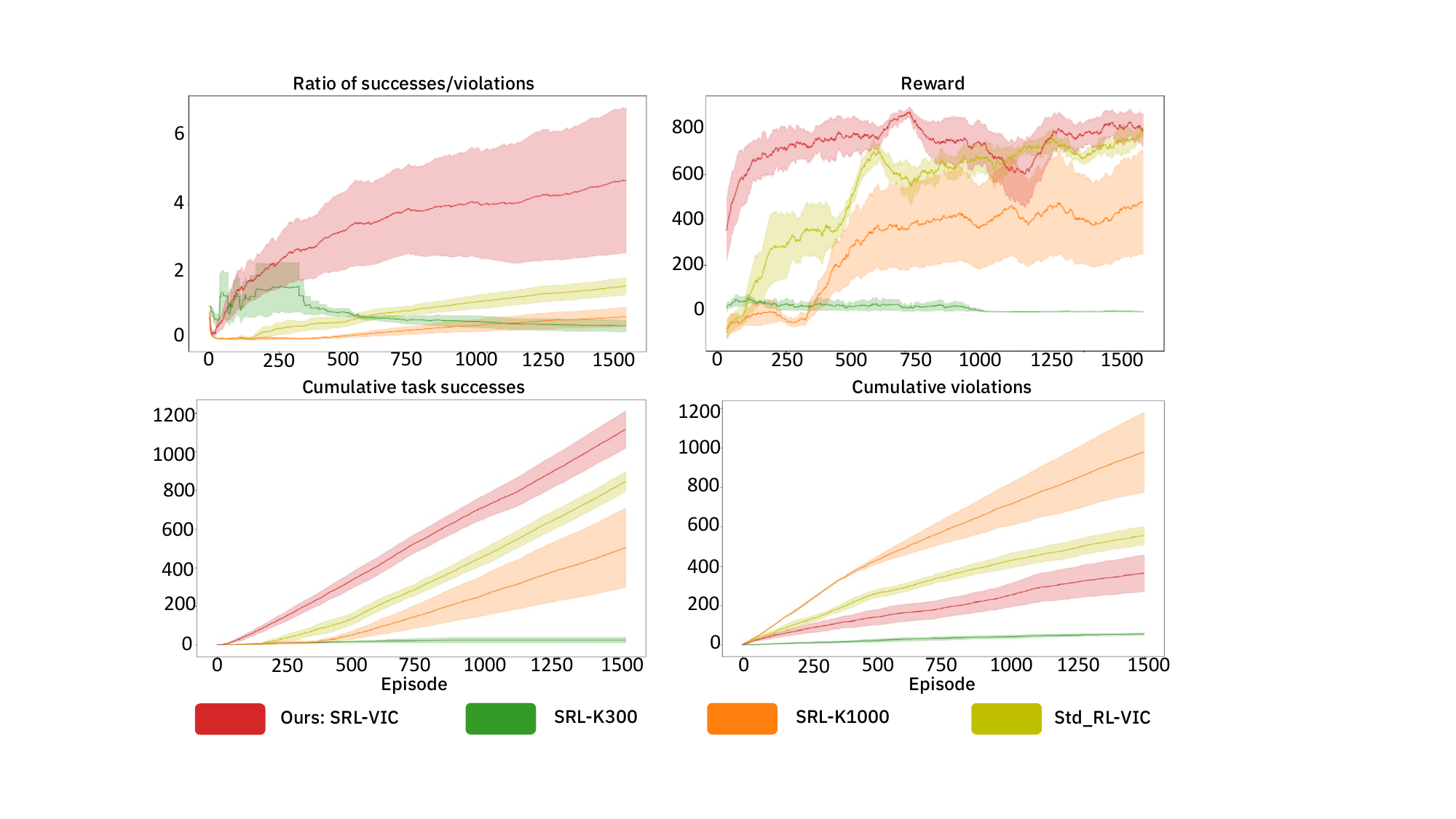}
    \vspace{-0.2cm}
    \caption{Learning curves for maze exploration. Our framework (\texttt{SRL-VIC}) outperforms others, particularly in terms of ratio of successes/violation and cumulative violations, i.e., our framework is safer than others. Furthermore, it learns faster, achieving success at an earlier stage. The large deviation area in the ratio of successes/violations comes from the division operation; the cumulative successes and violations do not exhibit large variation.}
    \label{fig:training_analyse}
    \vspace{-0.3cm}
\end{figure}

\subsubsection{Online training} 
We train each setup online for 3 runs with 1500 episodes in each run. In each episode, the robot needs to move to the start point and then begin exploring the maze, until it meets termination conditions.
We train the policy using 1 chunk of 20 cores and one GPU (NVIDIA Tesla V100 16Gb) from the cluster Franklin High Performance Computing (HPC)\footnote[6]{\label{fn:hpc}We gratefully acknowledge the Data Science and Computation Facility and its Support Team at Fondazione Istituto Italiano di Tecnologia.}. The average training time for 3 runs of each experiment \texttt{SRL-VIC}, \texttt{Std\_RL-VIC}, \texttt{SRL-K1000}, \texttt{SRL-K300} is 14.47, 7.32, 9.08, 68.28 hours respectively. The time gets shorter when episodes terminate early by constraint violation or task success.

Fig.~\ref{fig:mulTraj} shows the evolution of the robot's behaviour with increasing episodes of online learning. 
In the beginning, the agent fails because it violates the constraints while interacting with the obstacles. The agent learns to complete the task in later episodes. 
The stiffness ellipsoids indicate that the agent learns to adapt the stiffness to different conditions such as obstacle or wall contact. 
We observe that the trajectory becomes smoother and the robot finishes the task faster with further training. 

\subsubsection{Results}
We compare the proposed framework with baselines in four different perspectives: ratio of successes/violations, reward, cumulative task successes, cumulative violations. Specifically, cumulative task successes indicate the effectiveness of task completion; while cumulative violations show the number of constraint violations that indicates the safety; ratio of successes/violations show the trade-off between task completion and safety guarantee. 
The results of the simulation training are presented in Fig.~\ref{fig:training_analyse}. 

In terms of safety, our method achieves minimal cumulative violations apart from \texttt{SRL-K300}. Although \texttt{SRL-K300} has the minimum cumulative violations among all setups, it gets stuck in the obstacle area and cannot move forward. 
Thus, our method has an overwhelming advantage in the ratio of successes/violations (Fig.~\ref{fig:training_analyse}). 
We also tested the final learned model of \texttt{SRL-VIC} 100 times in simulation and observed that it was successful in all cases without any violation.

\begin{figure}[tb] 
    \centering  
    \begin{overpic}[trim=0cm 0cm 0cm 3cm, width=0.94\columnwidth]{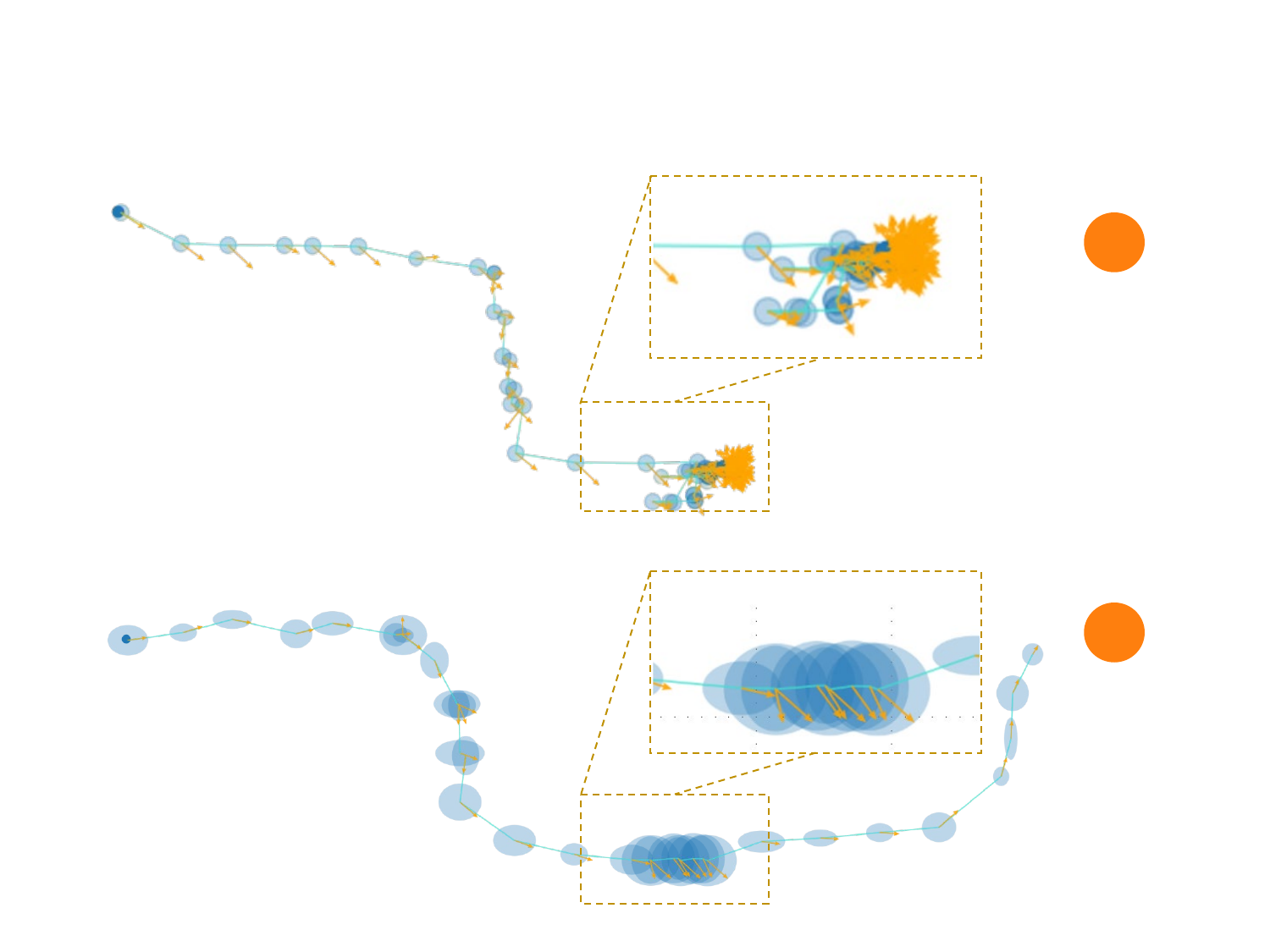}
        \put(6, 47){\small \bfseries \color{teal}{SRL-K300}}
        \put(6, 16){\small \bfseries \color{teal}{SRL-VIC}}
    \end{overpic}
      \vspace{-0.1cm}
      \caption[x]{Different behaviours when encountering obstacles at training episode 1000\textsuperscript{\ref{fn:traj}}. 
      Top (\texttt{SRL-K300}): The robot cannot move forward and fails due to low stiffness. Bottom (\texttt{SRL-VIC}): VIC approach switches to high stiffness when there are some obstacles on the way. The method learns to safely push the obstacles without exceeding the force threshold.}
      \label{fig:K300FailedwithObstacles}
      \vspace{-0.4cm}
\end{figure}

Our method achieves the best results also in means of task successes (Fig.~\ref{fig:training_analyse}). Although \texttt{Std\_RL-VIC} performed fairly well in means of reward and cumulative task successes, our framework improves it further. The difference is even stronger in the success/violation ratio results. 
Meanwhile, our method significantly outperforms the other \textit{recovery}-based methods (\texttt{SRL-K300} and \texttt{SRL-K1000}). \texttt{SRL-K300} cannot obtain high performance because it is not stiff enough to deal with obstacles. 
Fig.~\ref{fig:K300FailedwithObstacles} shows the advantage of our method over \texttt{SRL-K300} when encountering obstacles. Please note the increased size of the stiffness ellipsoids when the robot makes contact with the obstacles.

\begin{figure}[t] 
      \centering
      \begin{tabular}{c}
      \includegraphics[trim=1cm 0cm 0cm 0.5cm, width=0.96\columnwidth]{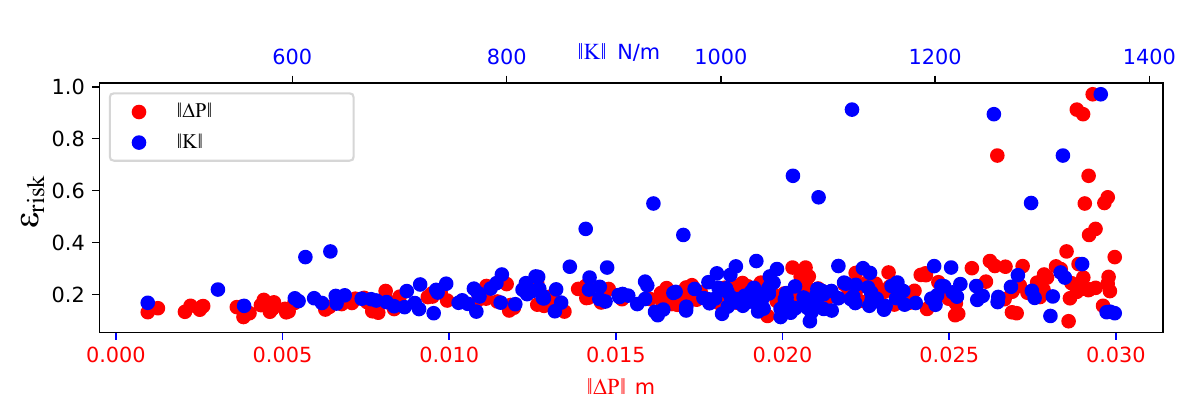}   
      \\          
      \includegraphics[trim=1cm 0.5cm 0cm 0.5cm, width=0.93\columnwidth]{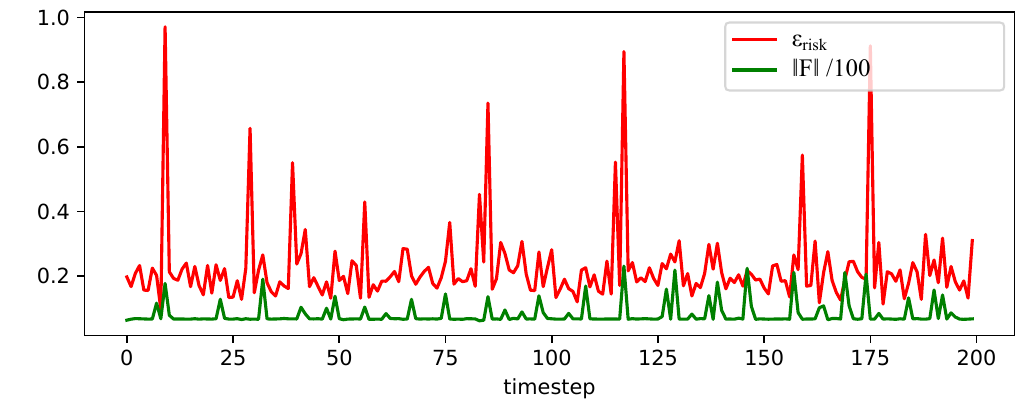} 
      \end{tabular}
      \vspace{-0.1cm}  
      \caption{ Relation between the risk value and state/action values in the offline test. 
      Top: The relationship of the risk $\epsilon_{risk}$ to stiffness $\left\| \boldsymbol{K}\right\|$ and desired move $\left\| \Delta\boldsymbol{P} \right\|$ values. Bottom: Plot of $\epsilon_{risk}$ and contact force $\left\| \boldsymbol{F} \right\|$. 
      The force magnitude is scaled down by $1/100$. 
       }
      \label{fig:model_test_all} 
      \vspace{-0.3cm}
\end{figure}

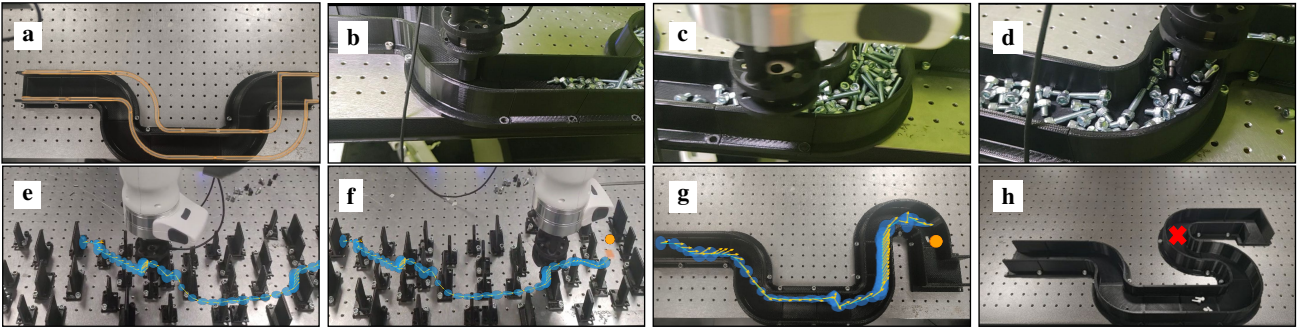
\begin{figure*}[tb] 
      \centering  
      \begin{overpic}[trim=0cm 9.5cm 0cm 1cm,width=0.97\linewidth]{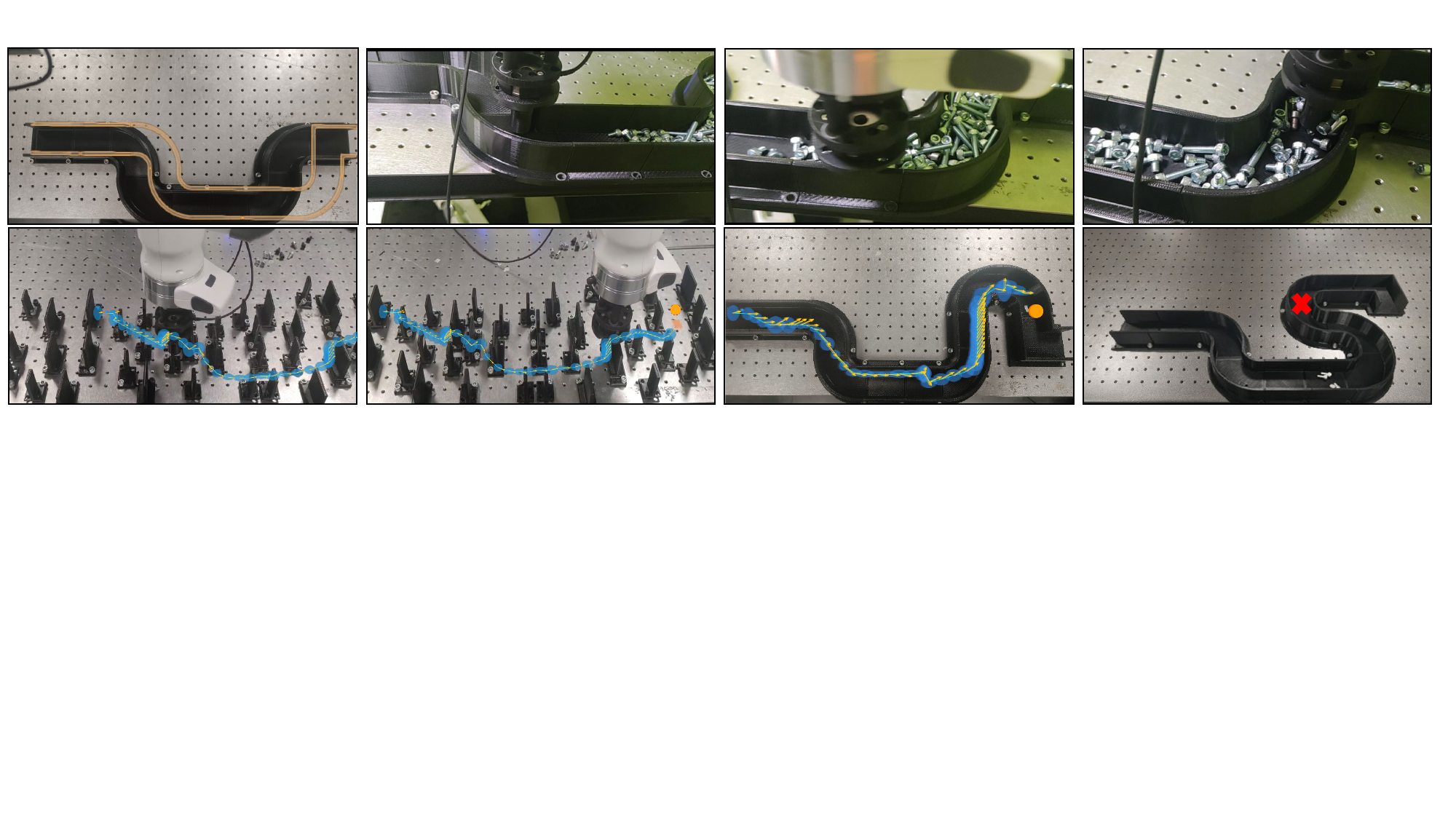} 
      \put(1.5, 22){\small \bfseries \colorbox{white}{a}}
      \put(26, 21.7){\small \bfseries \colorbox{white}{b}}
      \put(51, 22){\small \bfseries \colorbox{white}{c}}
      \put(75.5, 21.7){\small \bfseries \colorbox{white}{d}}
      \put(1.5, 10){\small \bfseries \colorbox{white}{e}}
      \put(26, 9.7){\small \bfseries \colorbox{white}{f}}
      \put(51, 10){\small \bfseries \colorbox{white}{g}}
      \put(75.5, 9.7){\small \bfseries \colorbox{white}{h}}
      \end{overpic}
        \vspace{-0.1cm}
      \caption{\textit{(a)} \textit{Maze shape-1}, without obstacles. The yellow overlay shows the original maze from the simulation.  \textit{(b)} The robot meets the pile of bolts halfway. \textit{(c)} The obstacles create congestion at the curve. \textit{(d)} The \texttt{SRL-VIC} policy overcomes the obstacles by increasing the stiffness. \textit{(e, f)} The \textit{complex task scenario} with small wall segments at new positions. \textit{(g)} \textit{Maze shape-2}. \textit{(h)}\textit{ Maze shape-3}, failed at red X marker.}
      \label{fig:real_exp_snapshot} 
      \vspace{-0.3cm}
\end{figure*} 

Regarding speed, \texttt{SRL-VIC} has to take relatively conservative actions in unsafe situations. Therefore, it achieved the task in slightly longer times compared to \texttt{SRL-K1000} and \texttt{Std\_RL-VIC}, but it was still faster than \texttt{K300}.

The results strongly demonstrate that our method provides a good trade-off between safety and task performance by combining the benefits of VIC and SRL. The fact that \texttt{Std\_RL-VIC} is the second-best method, performing better than the other baselines of SRL with constant stiffness values, confirms the importance of VIC in dealing with contact-rich tasks. 
The VIC allows switching between the low-stiffness behaviour and high-stiffness behaviour under different conditions to use the advantage of both. 
We also observe that it stays stiffer along the motion direction while it stays more compliant laterally.  
However, the proposed method improves the performance even further as the safety critic provides a predictive mechanism to avoid dangerous situations before getting too close. It decreases the violations, thus allows the task policy to explore more states without failing. 
Additionally, the recovery policy takes the robot away from the risky state, making it harder to get stuck.

\subsubsection{Safety critic behaviour}
In order to understand the behaviour of our \textit{safety critic} network, we visualise the relationship between the predicted risk value and the state/action values. 
The robot was commanded to repetitively apply a random action 200 times; starting on a random point (with 0.5 cm deviation), and moving in random direction (in [$0^\circ, 360^\circ$]) with random move size and stiffness.
We recorded the risk value produced for each of these cases and plotted them in Fig.~\ref{fig:model_test_all}.

As seen in the figure, higher $\Delta \boldsymbol{P}$ and $\boldsymbol{K}$ values are correlated with higher $\epsilon_{risk}$. This matches the human intuition that stronger actions are riskier. 
We also see that the model is sensitive to the contact force, as the the spikes in $\boldsymbol{F}$ are aligned with the spikes in $\epsilon_{risk}$. However, it does not merely follow the force, as there is a more complex relationship between the state/action values and risk. These outcomes confirm that the system acts reasonably. 

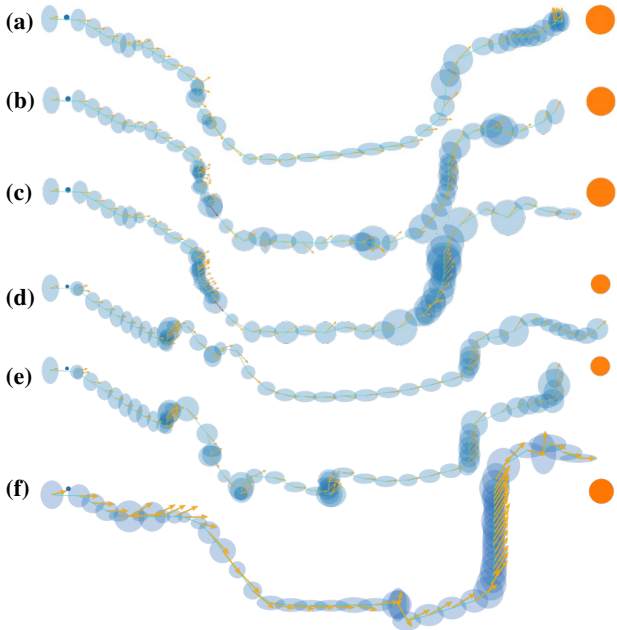
\begin{figure}[tb] 
      \centering 
      \begin{overpic}[trim=0.5cm 4cm 3cm 7.5cm, width=0.89\columnwidth]{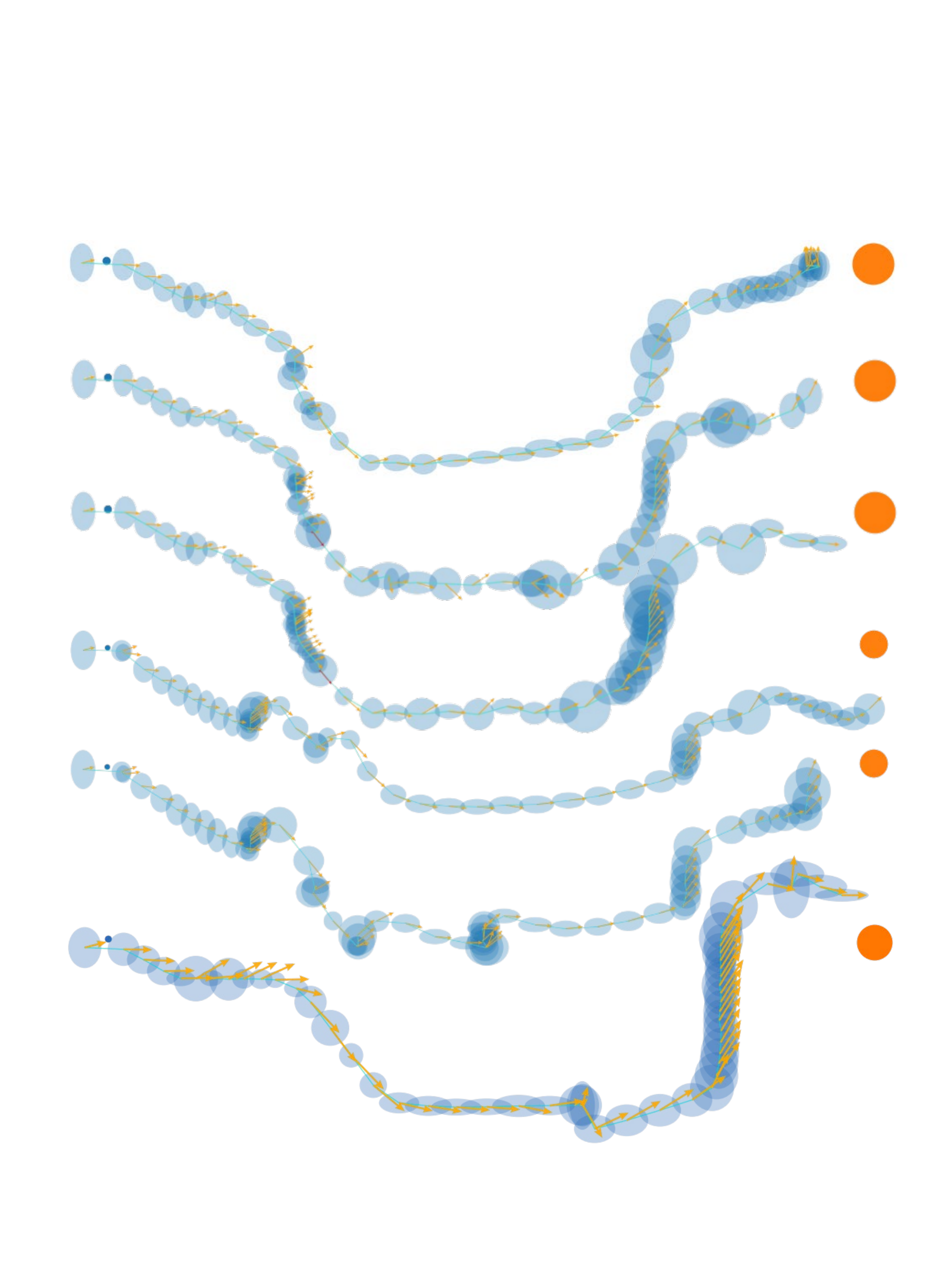}
      \put(0.5, 93){\small \bfseries \color{black}{(a)}}
      \put(0.5, 81){\small \bfseries \color{black}{(b)}}
      \put(0.5, 67){\small \bfseries \color{black}{(c)}}
      \put(0.5, 51){\small \bfseries \color{black}{(d)}}
      \put(0.5, 39){\small \bfseries \color{black}{(e)}}
      \put(0.5, 22){\small \bfseries \color{black}{(f)}}
      \end{overpic}
      \vspace{-0.1cm}
      \caption{Trajectories of the end-effector during executing task in the real world\textsuperscript{\ref{fn:traj}}. \textit{(a-c)} Maze shape-1 (Fig.\ref{fig:real_exp_snapshot}.a):
      \textit{(a)} \textit{no obstacle}, \textit{(b)} \textit{small obstacles}, \textit{(c)} \textit{mixed obstacles}.
      \textit{(d, e)} \textit{Complex task scenario} with discrete walls, \textit{(f)} \textit{Maze shape-2} (Fig.\ref{fig:real_exp_snapshot}.g).} 
      \label{fig:trajs_with_ellipses} 
      \vspace{-0.3cm}
\end{figure} 


\subsection{Real-world experiments}
We trained the policy in Mujoco for the contact-rich maze-exploration task and deployed it on a physical 7-DOF Franka robot arm without any fine-tuning. In the setup, we use Robot Operating System (ROS) where RL policy is a separate ROS node communicating with the physical robot.


Our policy succeeded in 4 out of 5 trials at first. In the failed case, the robot got stuck at a turning point. 
It was not highly robust to the real-world environment because our policy was trained with the observations in simulation. This is because of the physical dynamics differences and the imprecision of the F/T sensing in simulation. 
For this reason, we retrained the policy by adding Ornstein-Uhlenbeck(OU) noise in state observations. OU noise is a type of correlated noise often used in RL for continuous control tasks which helps the agent explore more systematically than purely random noise. The OU process is defined by a differential equation and has a mean-reverting property, which can be useful for stabilizing exploration. The success rate increased to 6/6 after retraining. Fig.~\ref{fig:real_exp_snapshot} shows the deployment on a physical robot. We put bolts of different size as dynamic obstacles (50 M6$\times$30 for \textit{small obstacles}, 100 M6$\times$8 and 50 M6$\times$30 for \textit{mixed obstacles}). A video of the experiments is available at \href{https://youtu.be/ksWXR3vByoQ}{https://youtu.be/ksWXR3vByoQ}. 

\begin{figure} 
\centering
\begin{tabular}{c}
\includegraphics[width=0.95\columnwidth]{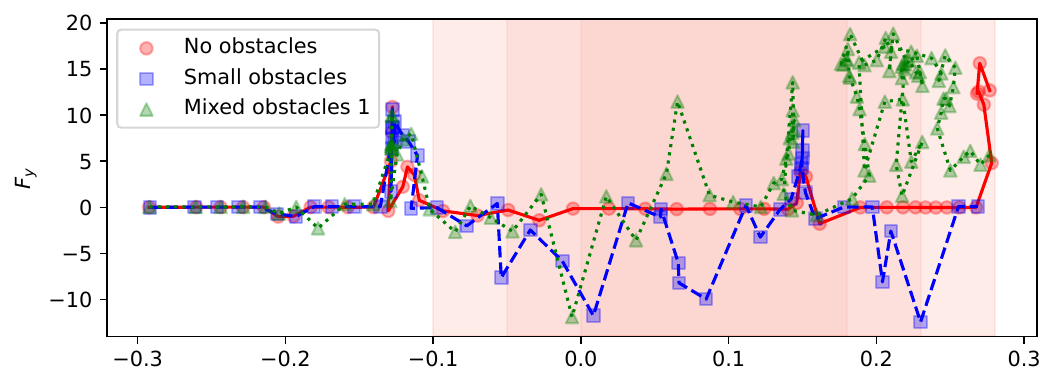} 
\vspace{-0.2cm}
\\
\includegraphics[width=0.95\columnwidth]{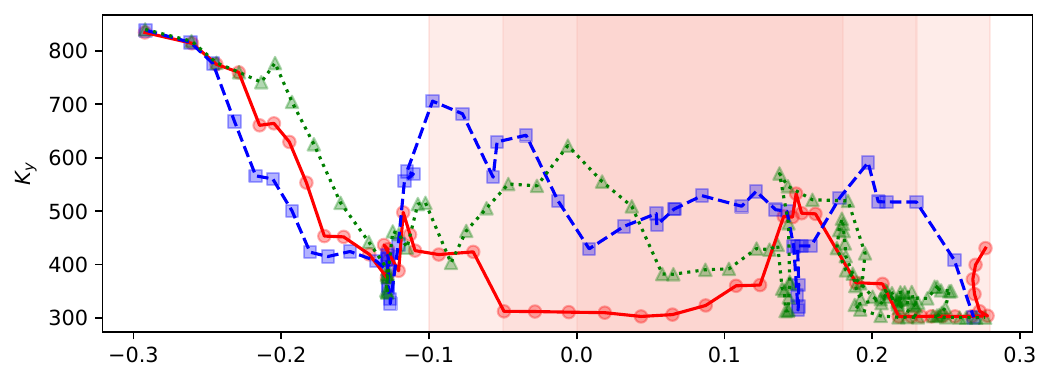} 
\vspace{-0.2cm}
\\       
\includegraphics[width=0.95\columnwidth]{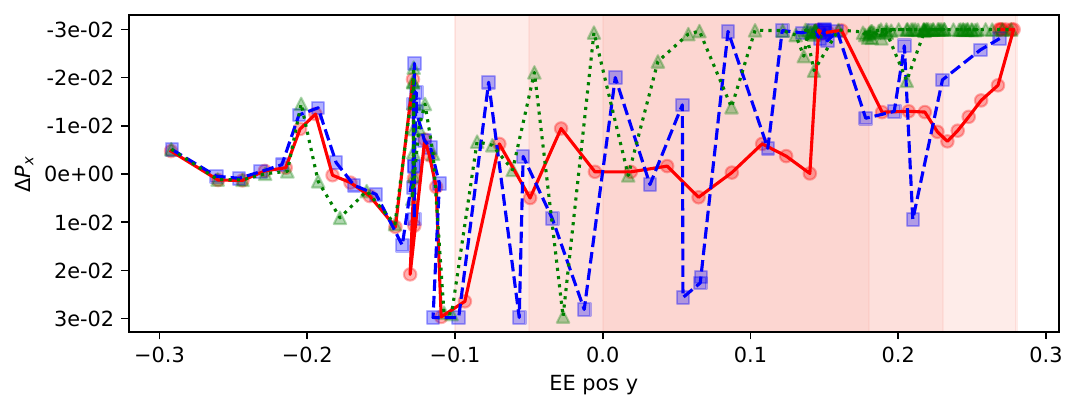}      
\end{tabular}
\vspace{-0.2cm}
\caption{Force ($F_y$), move size ($\Delta P_x$), and stiffness ($K_y$) values during the real-robot experiments plotted against the \textit{y-} position. The pink highlight shows the approximate placement of obstacles.}   
\label{fig:forceAndStiffness} 
\vspace{-0.3cm}
\end{figure}

\subsubsection*{Results} 
The \texttt{SRL-VIC} policy transfers successfully from simulation to real world. The transfer success was increased by the addition of state randomization in the simulation training. Apart from the physical dynamics and sensing differences between the simulation and the real world, there were also task-level differences, namely, the shape of the maze and the character of the obstacles.

In the real-world experiments, we tested the generalization capability of our safety model by changing the position of the maze curves and placing wall segments at new positions as shown in Fig.~\ref{fig:real_exp_snapshot}. 
The maze length varied between simulation (70.35 cm) and the real-world (84.62 cm). 
Additionally, the nature of the obstacles differed from those present in the simulation. 
The learned policy successfully completed most of these cases, suggesting that the safety mechanism generalizes to different wall positions. 
The agent can also complete the \textit{Maze shape-1} when tested by changing the flange
diameter (1, 2, 3 and 4 cm).
Our task policy is trained on a single maze shape; thus, it could not generalize to the case shown in Fig.~\ref{fig:real_exp_snapshot}.h. 
It still did not violate a constraint, but it could not find the exit.
Dealing with more diverse shapes of mazes requires training with more diverse shapes as well, however, we do not focus on task generalization in this work.

The end-effector trajectories in real-robot experiments are shown in Fig.~\ref{fig:trajs_with_ellipses}.
The policy adapts the stiffness values effectively by becoming stiff in the goal direction and staying compliant laterally. It also employs distinct strategies for encounters with walls at curves and collisions with movable obstacles. For the latter, it adopts a larger, rounded stiffness ellipse to facilitate pushing through in all directions.

We also plot the measured force ($F_y$), stiffness $K_y$ and move size ($\Delta P_x$) values w.r.t. \textit{y-}position in Fig.~\ref{fig:forceAndStiffness}, as \textit{y} is the main axis towards the goal. 
The stiffness adaptation policy becomes more evident in these plots, comparing the \textit{no obstacle} case to the others. 
The $2^{nd}$ and $3^{rd}$ cases have larger stiffness in the obstacle area as they observe higher force, while the $1^{st}$ case increases stiffness only at the curves, while touching the wall. Another difference is seen in the $\Delta P_x$ action, as the $2^{nd}$ and $3^{rd}$ cases move laterally to open the way through the obstacles, while the $1^{st}$ maintains its \textit{x-}position.

\section{Conclusion and future work}

We presented a framework that integrates the strengths of a safe RL method with VIC. This combination addresses the dual challenges of ensuring safety and maintaining continuous physical contact. 
Experiments in contact-rich maze tasks demonstrate that our framework achieved a good trade-off between task accomplishment and collision-based safety, and outperformed \texttt{Std\_RL-VIC}. The inclusion of the VIC adds increased adaptability, creating a better trade-off between safety and success in comparison to the constant impedance approaches.
The trained model can transfer to real-world, and also the safety model can generalize to different obstacles and wall positions. 
In the future, we will evaluate our framework in more generalized task scenarios and introduce model-based RL into our framework to improve the safety and robustness of our proposed framework in unknown environments.



\bibliographystyle{IEEEtran}
\bibliography{root}

\end{document}